\documentclass[runningheads,a4paper]{llncs}

\usepackage{amsmath,epsfig}
\usepackage{url}
\usepackage{multirow}
\usepackage{rotating}
\usepackage{amssymb}
\usepackage{graphicx,amsfonts}
\usepackage{algorithm,algorithmic}
\newcommand{\keywords}[1]{\par\addvspace\baselineskip
\noindent\keywordname\enspace\ignorespaces#1}

\newcommand{\cH}{\mathcal{H}}

\newcommand{\cL}{\mathcal{L}}

\newcommand{\cC}{\mathcal{C}}

\newcommand{\bZ}{\boldsymbol{Z}}

\newcommand{\bT}{\boldsymbol{T}}

\newcommand{\bx}{\boldsymbol{x}}
\newcommand{\bh}{\boldsymbol{h}}
\newcommand{\by}{\boldsymbol{y}}

\newcommand{\bz}{\boldsymbol{z}}
\newcommand{\ba}{\boldsymbol{a}}

\newcommand{\bc}{\boldsymbol{c}}

\newcommand{\bw}{\boldsymbol{w}}

\newcommand{\bbf}{\boldsymbol{f}}

\newcommand{\bZero}{\boldsymbol{0}}

\newcommand{\beq}{\begin{equation}}
\newcommand{\eeq}{\end{equation}}
\newcommand{\beqn}{\begin{eqnarray}}
\newcommand{\eeqn}{\end{eqnarray}}
\newcommand{\beqns}{\begin{eqnarray*}}
\newcommand{\eeqns}{\end{eqnarray*}}

\newcommand{\R}{\mathbb{R}}
\newcommand{\HH}{\mathbb{H}}

\newcommand{\C}{\mathbb{C}}

\newcommand{\N}{\mathbb{N}}

\newcommand{\frechet}{\textrm{Fr\'{e}chet }}
\newcommand{\fredif}{\textrm{Fr\'{e}chet differentiable }}

\makeatletter
\newcommand{\bdiv}{\mathop{\operator@font div}}
\makeatother

\makeatletter
\newcommand{\diag}{\mathop{\operator@font diag}}
\makeatother

\makeatletter
\newcommand{\conv}{\mathop{\operator@font conv}}
\makeatother

\makeatletter
\newcommand{\sign}{\mathop{\operator@font sign}}
\makeatother \makeatletter

\makeatletter
\newcommand{\proj}{\mathop{\operator@font proj}}
\makeatother \makeatletter

\makeatletter
\newcommand{\spa}{\mathop{\operator@font span}}
\makeatother \makeatletter

\makeatletter
\newcommand{\epi}{\mathop{\operator@font epi}}
\makeatother \makeatletter

\makeatletter
\newcommand{\dom}{\mathop{\operator@font dom}}
\makeatother \makeatletter

\begin{document}

\mainmatter  

\title{The Complex Gaussian Kernel LMS algorithm}

\titlerunning{Complex Gaussian Kernel LMS}

\author{Pantelis Bouboulis \and Sergios Theodoridis}

\authorrunning{P. Bouboulis and S. Theodoridis}


\institute{Department of Informatics and Telecommunications,\\
University of Athens, Athens, Greece.\\
\{bouboulis, stheodor\}@di.uoa.gr
}

%
%

\toctitle{Complex Gaussian Kernel LMS}
\tocauthor{Bouboulis - Theodoridis}
\maketitle

\begin{abstract}
Although the real reproducing kernels are used in an increasing number of machine learning problems, complex kernels have not, yet, been used, in spite of their potential interest in applications such as communications. In this work, we focus our attention on the complex gaussian kernel and its possible application in the complex Kernel LMS algorithm. In order to derive the gradients needed to develop the complex kernel LMS (CKLMS), we employ the powerful tool of Wirtinger's Calculus, which has recently attracted much attention in the signal processing community. Writinger's calculus simplifies computations and offers an elegant tool for treating complex signals. To this end, the notion of Writinger's calculus is extended to include complex RKHSs. Experiments verify that the CKLMS offers significant performance improvements over the traditional complex LMS or Widely Linear complex LMS (WL-LMS) algorithms, when dealing with nonlinearities.
\end{abstract}

\section{Introduction}

In recent years, kernel based algorithms have become the state of the art for many problems, especially in the machine learning community. The common feature of these problems is that they are casted as optimization problems over a Reproducing Kernel Hilbert Space (RKHS). The main advantage of  mobilizing the tool of RKHSs is that the original nonlinear task is ``transformed'' into a linear one, where one can employ an easier ``algebra".  Moreover, different types of nonlinearities can be treated in a unifying way, that does not affect  the derivation of the algorithms, except at the final implementation stage. The main concepts of this procedure can be summarized in the following two steps: 1) Map the finite dimensionality input data from the input space $F$ (usually $F\subset \R^\nu$) into a higher dimensionality (possibly infinite) RKHS $\cH$ and 2) Perform a linear processing (e.g., adaptive filtering) on the mapped data in $\cH$. The procedure is equivalent with a non-linear processing (non-linear filtering) in $F$.

An alternative way of describing this process is through the popular \textit{kernel trick} \cite{SchoSmo}, \cite{TheoKou}: ``Given an algorithm, which is formulated in terms of dot products, one can construct an alternative algorithm by replacing each one of the dot products with a positive definite kernel $\kappa$''.  The specific choice of kernel, implicitly, defines  a RKHS with an appropriate inner product. Furthermore, the choice of a kernel also defines the type of nonlinearity that underlies the model to be used. Although there are several kernels available in the relative literature, in most cases the powerful real Gaussian kernel is adopted.

The main representatives of this class of algorithms are the celebrated \textit{support vector machines} (SVMs), which have dominated the research in machine learning over the last decade. Moreover, processing in Reproducing Kernel Hilbert Spaces (RKHSs) in the context of online adaptive processing is also gaining in popularity within the signal processing community \cite{LiPokPrin}, \cite{KivSmoWil}, \cite{EngManMe}, \cite{SlaTheYam}, \cite{SlaTheYam2}. Besides SVMs and the more recent applications in adaptive filtering, there is a plethora of other scientific domains that have gained from adopting kernel methods (e.g., image processing and denoising \cite{KimFraScho}, \cite{BouSlaThe}, principal component analysis \cite{SchoSmoMu}, clustering \cite{FilCaMaRo}, e.t.c.).

Although the real Gaussian RBF kernel is quite popular in the aforementioned context, the existence of the corresponding complex Gaussian kernel is relatively unknown to the machine learning community. This is partly due to the fact, that in classification tasks (which is the dominant application of kernel methods)  the use of complex kernels is prohibitive, since no arrangement can be derived in complex domains and the necessary separating hyperplane of SVMs cannot be defined. Consequently, all known kernel based applications, since they emerged from the specific background, use real-valued kernels and they are able to deal with real valued data sequences only. While the complex gaussian RBF kernel is known to the mathematicians (especially those working on Reproducing Kernel Hilbert Spaces or Functional Analysis), it has remained in obscurity in the machine learning society.  In this paper, however, we use the complex gaussian kernel to address the problem of adaptive filtering of complex signals in RKHSs, focusing on the recently developed Kernel LMS (KLMS) \cite{LiPokPrin}, \cite{LiuPriHay}. The main goals of this paper are: a) to elevate from obscurity the complex Gaussian kernel as an effective tool for kernel based adaptive processing of complex signals, b) the extension of \textit{Wirtinger's Calculus} in complex RKHSs as a means for the elegant and efficient computation of the gradients, that are involved in many adaptive filtering algorithms, and c) the development of the Complex Kernel LMS (CKLMS) algorithm, by exploiting the extension of Wirtinger's calculus and the RKHS of complex gaussian kernels.  Wirtinger's calculus \cite{Wirti} is enjoying  increasing popularity, recently, mainly in the context of \textit{Widely Linear} complex adaptive filters \cite{Picin95}, \cite{Adali10}, \cite{Adali08a}, \cite{MatPaSte}, \cite{Moreno}, providing a tool for the derivation of gradients in the complex domain.

The paper is organized as follows. In section \ref{SEC:PRELIM} we provide a minimal introduction to complex RKHSs focusing on the complex gaussian kernel and its relation with the real one. Next, in section \ref{SEC:Wirtinger} we summarize the main notions of the extended Wirtinger's Calculus. Section \ref{SEC:CKLMS} presents the gaussian complex kernel LMS algorithm. Finally, experimental results and conclusions are provided in Section \ref{SEC:Experim}. We will denote the set of all real and complex numbers by $\R$ and $\C$ respectively. Vector or matrix valued quantities appear in boldfaced symbols.

\section{Reproducing Kernel Hilbert Spaces}\label{SEC:PRELIM}
In this section we briefly describe the Reproducing Kernel Hilbert Spaces. Since we are mainly interested on the complex case, we recall the basic facts on RKHS associated with complex kernels. The material presented here may be found with more details in \cite{Saitoh} and \cite{Paulsen}.
Given a function $\kappa:X\times X\rightarrow\C$ and $x_1,\dots,x_N \in X$, the matrix\footnote{The term $(K_{i,j})^N$ denotes a square $N\times N$ matrix.} $K=(K_{i,j})^N$ with elements $K_{i,j}=\kappa(x_i,x_j)$, for $i,j=1,\dots,N$, is called the \textit{Gram matrix} (or \textit{kernel matrix}) of $\kappa$ with respect to $x_1,\dots,x_N$.
A complex hermitian matrix $K=(K_{i,j})^N$ satisfying
\begin{align*}
c^H\cdot K\cdot c=\sum_{i=1,j=1}^{N,N} c_i^* c_j K_{i,j}\geq 0,
 \end{align*}
for all $c_i\in\C$, $i=1,\dots,N$, is called \textit{Positive Definite}\footnote{
In matrix analysis literature, this is the definition of the positive semidefinite matrix.}. Let $X$ be a nonempty set. Then a function $\kappa:X\times X\rightarrow\C$, which for all $N\in\N$ and all $x_1,\dots,x_N\in X$ gives rise to a positive definite Gram matrix $K$ is called a \textit{Positive Definite Kernel}.
In the following we will frequently refer to a positive definite kernel simply as \textit{kernel}.

Next, consider a linear class $\cH$ of complex valued functions $f$ defined on a set
$X$. Suppose further, that in $\cH$ we can define an inner product
$\langle\cdot,\cdot\rangle_\cH$ with corresponding norm
$\|\cdot\|_\cH$ and that $\cH$ is complete with respect to that
norm, i.e., $\cH$ is a Hilbert space. We call $\cH$ a
\textit{Reproducing Kernel Hilbert Space (RKHS)}, if for all $x\in
X$ the evaluation functional $T_x:\cH\rightarrow\C:\;T_x(f)=f(x)$ is
a continuous (or, equivalently, bounded) operator. If this is true,
then by the Riesz's representation theorem, for all $x\in X$ there
is a function $g_x\in\cH$ such that $T_x(f)=f(x)=\langle f,
g_x\rangle_\cH$. The function $\kappa:X\times
X\rightarrow\C:\;\kappa(y,x)=g_x(y)$ is called a \textit{reproducing
kernel} of $\cH$. It can be easily proved that the function $\kappa$
is a positive definite kernel.

Alternatively, we can define a RKHS as a Hilbert space $\cH$ for which there exists a function $\kappa:X\times X\rightarrow\C$ with the following two properties:
\begin{enumerate}
\item For every $x\in X$, $\kappa(\cdot,x)$ belongs to $\cH$.
\item $\kappa$ has the so called \textit{reproducing property}, i.e.
\begin{align}\label{EQ:rep_prop}
f(x)=\langle f,\kappa(\cdot, x)\rangle_\cH, \textrm{ for all } f\in\cH,
\end{align}
in particular $\kappa(x,y)=\langle \kappa(\cdot, y), \kappa(\cdot, x)\rangle_\cH$.
\end{enumerate}

It has been proved (see \cite{Aron50}) that to every positive definite kernel $\kappa$ there corresponds one and only one class of functions $\cH$ with a uniquely determined inner product in it, forming a Hilbert space and admitting $\kappa$ as a reproducing kernel. In fact the kernel $\kappa$ produces the entire space $\cH$, i.e.,
$\cH=\overline{\spa\{\kappa(x,\cdot)|x\in X\}}.$
The map $\Phi:X\rightarrow\cH:\Phi(x)=\kappa(\cdot,x)$ is called the \textit{feature map} of $\cH$.  Recall, that in the case of complex Hilbert spaces the inner product is sesqui-linear and Hermitian. In the real case the condition $\kappa(x,y)=\langle \kappa(\cdot, y), \kappa(\cdot, x)\rangle_\cH$ may be replaced by the well known equation $\kappa(x,y)=\langle \kappa(\cdot, x), \kappa(\cdot, y)\rangle_\cH$. However, since in the complex case the inner product is Hermitian, the aforementioned condition is equivalent to $\kappa(x,y)=\left(\langle \kappa(\cdot, x), \kappa(\cdot, y)\rangle_\cH\right)^*$.

Consider the complex valued function
\begin{align}\label{EQ:complex_gaussian_kernel}
\kappa_{\sigma,\C^d}(\bz,\bw) : = \exp\left(-\frac{\sum_{i=1}^{d}(z_i-w_i^*)^2}{\sigma^2}\right),
\end{align}
defined on $\C^d\times\C^d$, where $\bz,\bw\in\C^d$, $z_i$ denotes the $i$-th component of the complex vector $\bz\in\C^d$ and $\exp$ is the extended exponential function in the complex domain. It can be shown that $\kappa_{\sigma,\C^d}$ is a $\C$-valued kernel on $\C^d$, which we call the \textit{complex Gaussian kernel} with parameter $\sigma$. Its restriction $\kappa_{\sigma}:=\left(\kappa_{\sigma,\C^d}\right)_{|\R^d\times\R^d}$ is the well known \textit{real Gaussian kernel}:
\begin{align}\label{EQ:real_gaussian_kernel}
\kappa_{\sigma,\R^d}(\bx,\by) : = \exp\left(-\frac{\sum_{i=1}^{d}(x_i-y_i)^2}{\sigma^2}\right).
\end{align}
An explicit description of the RKHSs of these kernels, together with some important properties can be found in \cite{SteinHuSco}.

\section{Wirtinger's Calculus in complex RKHS}\label{SEC:Wirtinger}

Wirtinger's calculus \cite{Wirti} has become very popular in the signal processing community mainly in the context of
complex adaptive filtering \cite{Picin95}, \cite{ManGoh}, \cite{Adali10}, \cite{Adali08a},  \cite{CaGePaVe}, as a means of computing, in an elegant way,  gradients of real valued cost functions defined on complex domains ($\C^\nu$). The Cauchy-Riemann conditions dictate that such functions are not holomorphic and therefore the complex derivative cannot be used. Instead, if we consider that the cost function is defined on a Euclidean domain with a double dimensionality ($\R^{2\nu}$), then the real derivatives may be employed. The price of this approach is that the computations become cumbersome and tedious. Wirtinger's calculus provides an alternative equivalent formulation, that is based on simple rules and principles and which bears a great resemblance to the rules of the standard complex derivative. A self-consistent presentation of the main ideas of Wirtinger's calculus may be found in the excellent and highly recommended introductory report of K. Kreutz-Delgado \cite{Delga}.

In the case of a simple non-holomorphic complex function $T$ defined on $U\subseteq\C$, Wirtinger's calculus considers two forms of derivatives, the \textit{$\R$-derivative} and the \textit{conjugate $\R$-derivative}, which are defined as follows:
\begin{align*}
\frac{\partial T}{\partial z}=\frac{1}{2}\left(\frac{\partial u}{\partial x}+\frac{\partial v}{\partial y}\right) + \frac{i}{2}\left(\frac{\partial v}{\partial x}-\frac{\partial u}{\partial y}\right),\\
\frac{\partial T}{\partial z^*}=\frac{1}{2}\left(\frac{\partial u}{\partial x}-\frac{\partial v}{\partial y}\right) + \frac{i}{2}\left(\frac{\partial v}{\partial x}+\frac{\partial u}{\partial y}\right)
\end{align*}
where $T(z)=T(x+iy)=T(x,y)=u(x,y)+i v(x,y)$. Note that any such non-holomorphic function can be written in the form $T(z,z^*)$. Having this in mind, $\frac{\partial T}{\partial z}$,  can be easily evaluated as the standard complex partial derivative taken with respect to $z$ (thus treating $z^*$ as a constant). Consequently,  $\frac{\partial T}{\partial z^*}$ is evaluated as the standard complex partial derivative taken with respect to $z^*$ (thus treating $z$ as a constant). For example, if $T(z,z^*)=z(z^*)^2$, then $\frac{\partial T}{\partial z} = (z^*)^2,\quad \frac{\partial T}{\partial z^*}=2zz^*$.
Similar principles and rules hold for a function of many complex variables (i.e., $U\subseteq\C^{\nu}$) \cite{Delga}.

Wirtinger's calculus has been developed only for operators defined on finite dimensional spaces, $\C^{\nu}$. Hence, this calculus cannot be used in RKH spaces, where the dimensionality of the function space can be infinite, as, for example, it is the case for the Gaussian RKHSs. To this end, Wirtinger's calculus needs to be generalized to a general Hilbert space. A rigorous presentation of this extension is out of the scope of the paper (due to lack of space). Nevertheless, we will present the main ideas and results. We employ the \frechet derivative, a notion that generalizes differentiability on abstract Banach or Hilbert spaces. Consider a Hilbert space $H$ over the field $F$  (typically $\R$ or $\C$). The operator $T:H\rightarrow F$ is said to be \textit{\fredif} at $f_0$, if there exists a $u\in H$, such that
\begin{align}\label{EQ:frechet2}
\lim_{\|h\|_{H}\rightarrow 0}\frac{T(f_0+h)-T(f_0)-\langle u, h\rangle_{H}}{\|h\|_{H}}=0,
\end{align}
where $\langle\cdot, \cdot\rangle_{H}$ is the dot product of the Hilbert space $H$ and $\|\cdot\|_H=\sqrt{\langle\cdot, \cdot\rangle_H}$ is the
induced norm. The element $u$ is usually called the gradient of $T$ at $f_0$.

Assume that $\bT=(T_1,T_2)^T$, $\bT(\bbf) = \bT(f_1+i f_2) = \bT(f_1,f_2) = T_1(f_1,f_2) + i T_2(f_1,f_2)$, is differentiable as an operator defined on the RKHS $\cH$ and let $\nabla_{1}T_1$, $\nabla_{2}T_1$, $\nabla_{1}T_2$ and $\nabla_{2}T_2$ be the partial derivatives, with respect to the first ($f_1$) and the second ($f_2$) variable respectively. It turns out, proofs are omitted due to lack of space, that if $\bT(f_1,f_2)$ has derivatives of any order, then it can be written in
the form $\bT(\bbf, \bbf^*)$, where $\bbf^*=f_1-i f_2$, so that for fixed $\bbf^*$, $\bT$ is $\bbf$-holomorphic and for
fixed $\bbf$, $T$ is $\bbf^*$-holomorphic. We may define the $\R$-derivative and the conjugate $\R$-derivative of $\bT$ as follows:
\begin{align}
\nabla_{\bbf}\bT &= \frac{1}{2}\left(\nabla_1 T_1 + \nabla_2 T_2\right) + \frac{i}{2}\left(\nabla_1 T_2 - \nabla_2 T_1\right)\\
\nabla_{\bbf^*}\bT &= \frac{1}{2}\left(\nabla_1 T_1 - \nabla_2 T_2\right) + \frac{i}{2}\left(\nabla_1 T_2 + \nabla_2 T_1\right).
\end{align}
The following properties can be proved (among others):
\begin{enumerate}
\item The first order Taylor expansion around $\bbf\in\cH$ is given by
\begin{align*}
\bT(\bbf+\bh) =& \bT(\bbf) + \langle \bh, \left(\nabla_{\bbf} \bT(\bbf)\right)^* \rangle_\cH + \langle \bh^*, \left(\nabla_{\bbf^*} \bT(\bbf)\right)^* \rangle_\cH.
\end{align*}
\item If $\bT(\bbf)=\langle \bbf, \bw\rangle_\cH$, then $\nabla_{\bbf}\bT=\bw^*$, $\nabla_{\bbf^*}\bT=\bZero$.
\item If $\bT(\bbf)=\langle \bw, \bbf\rangle_\cH$, then $\nabla_{\bbf}\bT=\bZero$, $\nabla_{\bbf^*}\bT=\bw$.
\item If $\bT(\bbf)=\langle \bbf^*, \bw\rangle_\cH$, then $\nabla_{\bbf}\bT=\bZero$, $\nabla_{\bbf^*}\bT=\bw^*$.
\item If $\bT(\bbf)=\langle \bw, \bbf^*\rangle_\cH$, then $\nabla_{\bbf}\bT=\bw$, $\nabla_{\bbf^*}\bT=\bZero$.
\end{enumerate}

An important consequence of the above properties is that if $\bT$ is a real valued operator defined on $\cH$, then its first order Taylor's expansion is given by:
\begin{align*}
\bT(\bbf+\bh) & =  \bT(\bbf) + \langle \bh, \left(\nabla_{\bbf}\bT(\bbf)\right)^*\rangle_\cH + \langle \bh^*, \left(\nabla_{\bbf^*}\bT(\bbf)\right)^* \rangle_\cH\\
& = \bT(\bbf) + \langle \bh, \nabla_{\bbf^*}\bT(\bbf)\rangle_\cH + \left(\langle \bh, \nabla_{\bbf^*}\bT(\bbf)\rangle_\cH\right)^*\\
&= \bT(\bbf) + 2\cdot \Re\left[ \langle \bh, \nabla_{\bbf^*}\bT(\bbf)\rangle_\cH\right].
\end{align*}
However, in view of the Cauchy Riemann inequality we have:
\begin{align*}
\Re\left[ \langle \bh, \nabla_{\bbf^*}\bT(\bbf)\rangle_\cH\right] & \leq \left|\langle \bh, \nabla_{\bbf^*}\bT(\bbf)\rangle_\cH\right|
\leq \|\bh\|_\cH \cdot \| \nabla_{\bbf^*}\bT(\bbf)\|_\cH.
\end{align*}
The equality in the above relationship holds if $h\propto \nabla_{\bbf^*}\bT$. Hence, the direction of increase of $\bT$ is $\nabla_{\bbf^*}\bT(\bbf)$. Therefore, any gradient descent based algorithm minimizing $\bT(\bbf)$ is based on the update scheme:
\begin{align}
\bbf_{n} = \bbf_{n-1} - \mu\cdot\nabla_{\bbf^*}\bT(\bbf_{n-1}).
\end{align}

\section{Complex Kernel LMS}\label{SEC:CKLMS}
As an application of the complex gaussian kernel in adaptive filtering of complex signals, we focus on the recently developed \textit{Kernel Least Mean Squares Algorithm} (KLMS), which is the LMS algorithm in RKHSs \cite{LiPokPrin}, \cite{LiuPriHay}. KLMS, as all the known kernel methods that use real-valued kernels, was developed for real valued data sequences only. Here, the KLMS is extended to include the complex case. To our knowledge, no kernel-based strategy has been developed, so far, that is able to effectively deal with complex valued signals. Wirtinger's calculus is exploited to derive the necessary gradient updates.

Consider the sequence of examples $(\bz(1),d(1))$, $(\bz(2),d(2))$, $\dots$, $(\bz(N),d(N))$, where $d(n)\in\C$, $\bz(n)\in V\subset\C^\nu$, $\bz(n)=\bx(n) + i \by(n)$, $\bx(n), \by(n)\in\R^\nu$, for $n=1,\dots,N$. We map the points $\bz(n)$ to the gaussian complex RKHS $\cH$ using the feature map  $\Phi$,
for $n=1,\dots,N$. The objective of the complex Kernel LMS is to minimize $E\left[\cL_n(\bw)\right]$, where
\begin{align*}
\cL_n(\bw) &=  |e(n)|^2 = |d(n) - \langle \Phi(\bz(n)), \bw\rangle_\HH|^2\\
&= \left(d(n) - \langle \Phi(\bz(n)), \bw\rangle_\HH\right) \left(d(n) - \langle \Phi(\bz(n)), \bw\rangle_\HH\right)^*\\
&= \left(d(n) - \langle \bw^*, \Phi(\bz(n))\rangle_\HH\right) \left(d(n)^* - \langle \bw, \Phi(\bz(n))\rangle_\HH\right),
\end{align*}
at each instance $n$. We then apply the complex LMS to the transformed data, using the rules of Wirtinger's calculus to compute the gradient $\nabla_{\bw^*}\cL_n(\bw)=-e(n)^*\cdot\Phi(\bz(n))$.
Therefore the CKLMS update rule becomes
$\bw(n) = \bw(n-1) + \mu e(n)^*\cdot\Phi(\bz(n))$,
where $\bw(n)$ denotes the estimate at iteration $n$.

Assuming that $\bw(0)=\bZero$, the repeated application of the weight-update equation gives:
\begin{align}
\bw(n) =  \sum_{k=1}^{n} e(k)^*\Phi(\bz(k))\label{EQ:CKLMS_W}.
\end{align}
Thus, the filter output at iteration $n$ becomes:
\begin{align}
\hat d(n) =&\langle \Phi(\bz(n)), \bw(n-1) \rangle_\HH = \mu \sum_{k=1}^{n-1} e(k) \langle \Phi(\bz(n)), \Phi(\bz(k)) \rangle_\HH\nonumber\\
=& \mu \sum_{k=1}^{n-1} e(k) \kappa_{\sigma,\C^\nu}(\bz(n), \bz(k)).\nonumber
\end{align}

It can readily be shown that, since the CKLMS is the complex LMS in RKHS, the important properties of the LMS (convergence in the mean, misadjustment, e.t.c.) carry over to CKLMS. Furthermore, note that using the complex gaussian kernel the algorithm is automatically normalized.
The CKLMS algorithm is summarized in Algorithm \ref{ALG:NCKLMS1}. Although it is developed in the context of the complex gaussian kernel, it may be used with any other complex reproducing kernel.

\begin{algorithm}[h!]
\caption{Normalized Complex Kernel  LMS}\label{ALG:NCKLMS1}
\textbf{INPUT: } $(\bz(1),d(1))$, $\dots$, $(\bz(N),d(N))$\\
\textbf{OUTPUT:} The expansion \\
$\bw=\sum_{k=1}^{N} a(k)\kappa(\bz(k),\cdot)$.\\
\\
\textbf{Initialization:} Set  $\ba=\{\}$, $\bZ=\{\}$ (i.e., $\bw=\bZero$). Select the step parameter $\mu$ and the parameter $\sigma$ of the complex gaussian kernel.
\begin{algorithmic}
\FOR{n=1:N}
\STATE{Compute the filter output: $\hat d(n) = \sum_{k=1}^{n-1}a(k)\cdot\kappa_{\sigma,\C^\nu}(\bz(n),\bz(k))$.}
\STATE{Compute the error: $e(n)=d(n)-\hat d(n)$.}
\STATE{$a(n)=\mu e(n)$.}
\STATE{Add the new center $\bz(n)$ to the list of centers, i.e., add $\bz(n)$ to the list $\bZ$, add $a(n)$ to the list $\ba$.}
\ENDFOR
\end{algorithmic}
\end{algorithm}

In CKLMS,  we start from an empty set (usually called the \textit{dictionary}) and gradually add new samples to that set, to form a summation similar to the one shown in equation (\ref{EQ:CKLMS_W}). This results to an increasing memory and computational requirements, as time evolves. To cope with this problem and to produce sparse solutions, we employ the well known \textit{novelty criterion} \cite{Platt}, \cite{LiuPriHay}. In novelty criterion online sparsification, whenever a new data pair $(\Phi(\bz_n),d_n)$ is considered, a decision is immediately made of whether to add the new center $\Phi(\bz_n)$ to the dictionary of centers $\cC$. The decision is reached following two simple rules. First, the distance of the new center $\Phi(\bz_n)$ from the current dictionary is evaluated: $dis = \min_{\bc_k\in\cC}\{\|\Phi(\bz_n) - \bc_k\|_\HH\}$.
If this distance is smaller than a given threshold $\delta_1$ (i.e., the new center is close to the existing dictionary), then the center is not added to $\cC$. Otherwise, we compute the prediction error $e_n = d_n - \hat d_n$. If $|e_n|$ is smaller than a predefined threshold $\delta_2$, then the new center is discarded. Only if $|e_n|\geq\delta_2$ the new center $\Phi(\bz_n)$ is added to the dictionary.

\begin{figure}
\begin{center}
\includegraphics[scale=0.35]{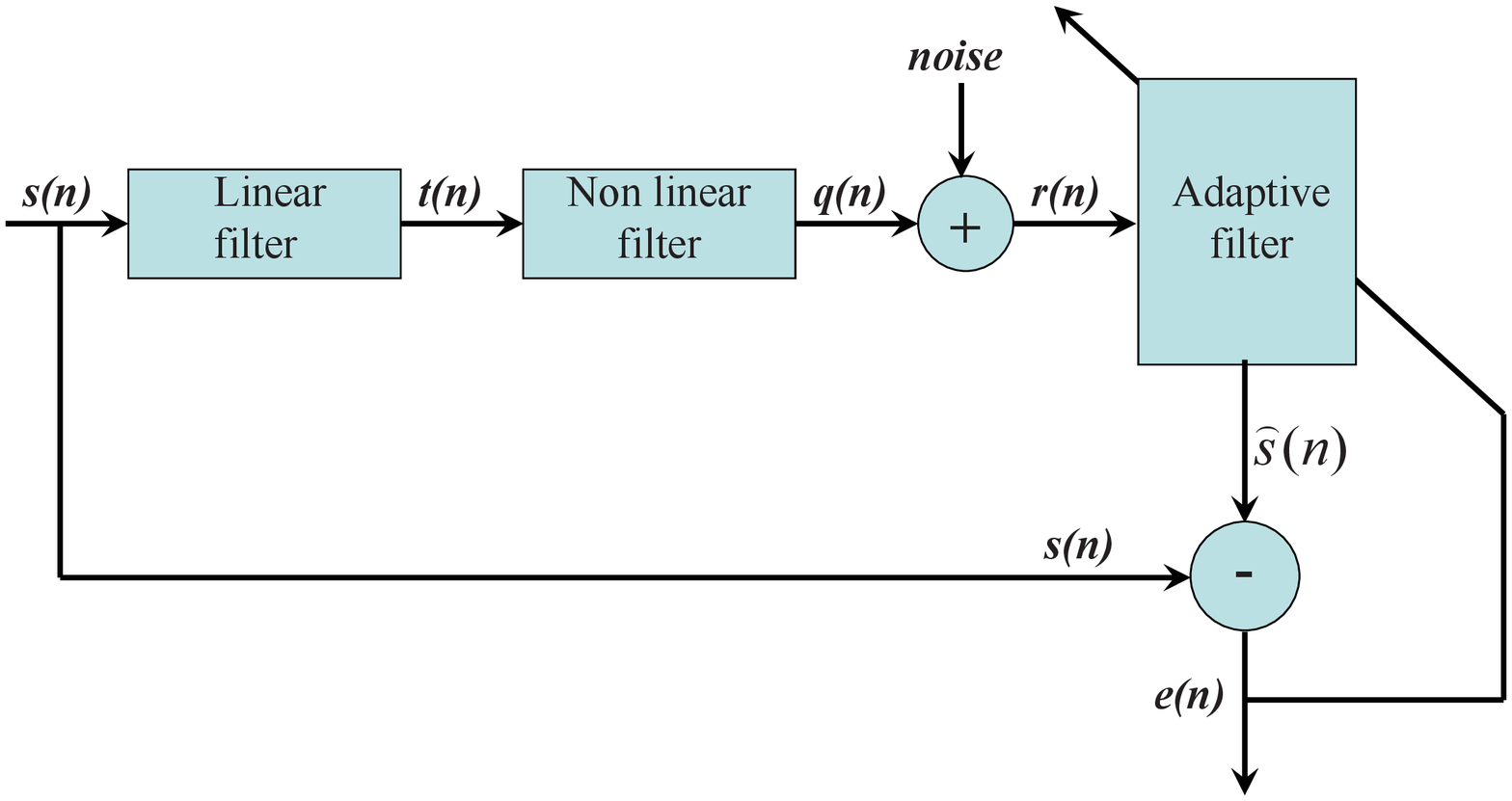}
\end{center}
\caption{The equalization problem.}\label{FIG:equal_form}
\end{figure}

\section{Experiments}\label{SEC:Experim}
We tested the CKLMS using  a simple nonlinear channel equalization problem (see figure \ref{FIG:equal_form}). The nonlinear channel consists of a linear filter: $t(n)= (-0.9+0.8i)\cdot s(n) + (0.6-0.7i)\cdot s(n-1)$
and a memoryless nonlinearity $q(n) = t(n) + (0.1+0.15i)\cdot t^2(n) + (0.06+0.05i)\cdot t^3(n)$.
At the receiver end of the channel, the signal is corrupted by white Gaussian noise and then observed as $r(n)$.
The input signal that was fed to the channel had the form
\begin{align}\label{EQ:input}
s(n) = 0.70(\sqrt{1-\rho^2}X(n) + i\rho Y(n)),
\end{align}
where $X(n)$ and $Y(n)$ are gaussian random variables. This input is circular for $\rho=\sqrt{2}/2$ and highly non-circular if $\rho$ approaches 0 or 1 \cite{Adali10}. The aim of channel equalization is to construct an inverse filter which taking the output $r(n)$, reproduces the original input signal with as low an error rate as possible. To this end we apply the NCKLMS algorithm to the set of samples
\begin{align*}
\left((r(n+D), r(n+D-1), \dots, r(n+D-L))^T, s(n)\right),
\end{align*}
where $L>0$ is the filter length and $D$ the equalization time delay.

\begin{figure}
\begin{center}
 \includegraphics[scale=0.3]{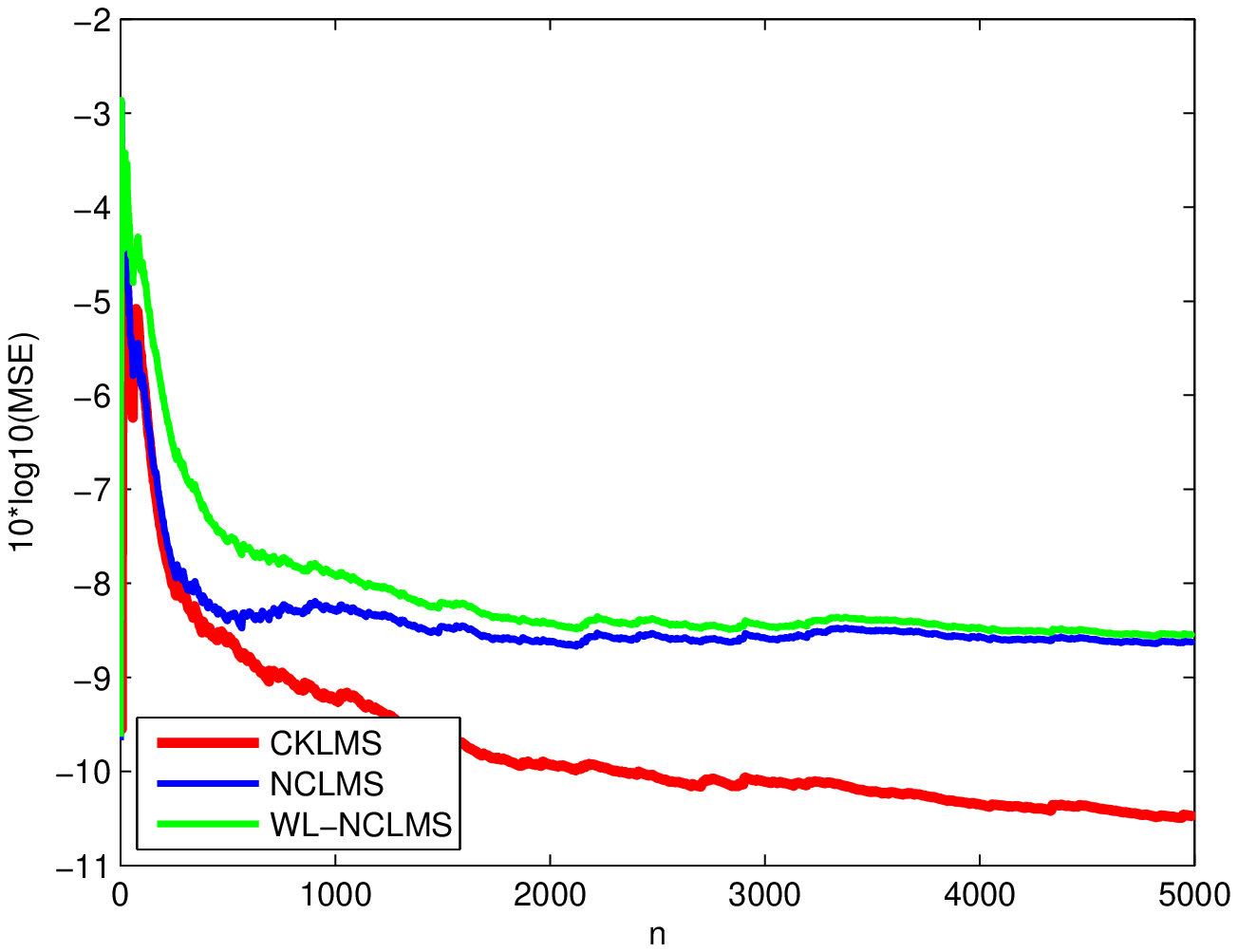}  \includegraphics[scale=0.3]{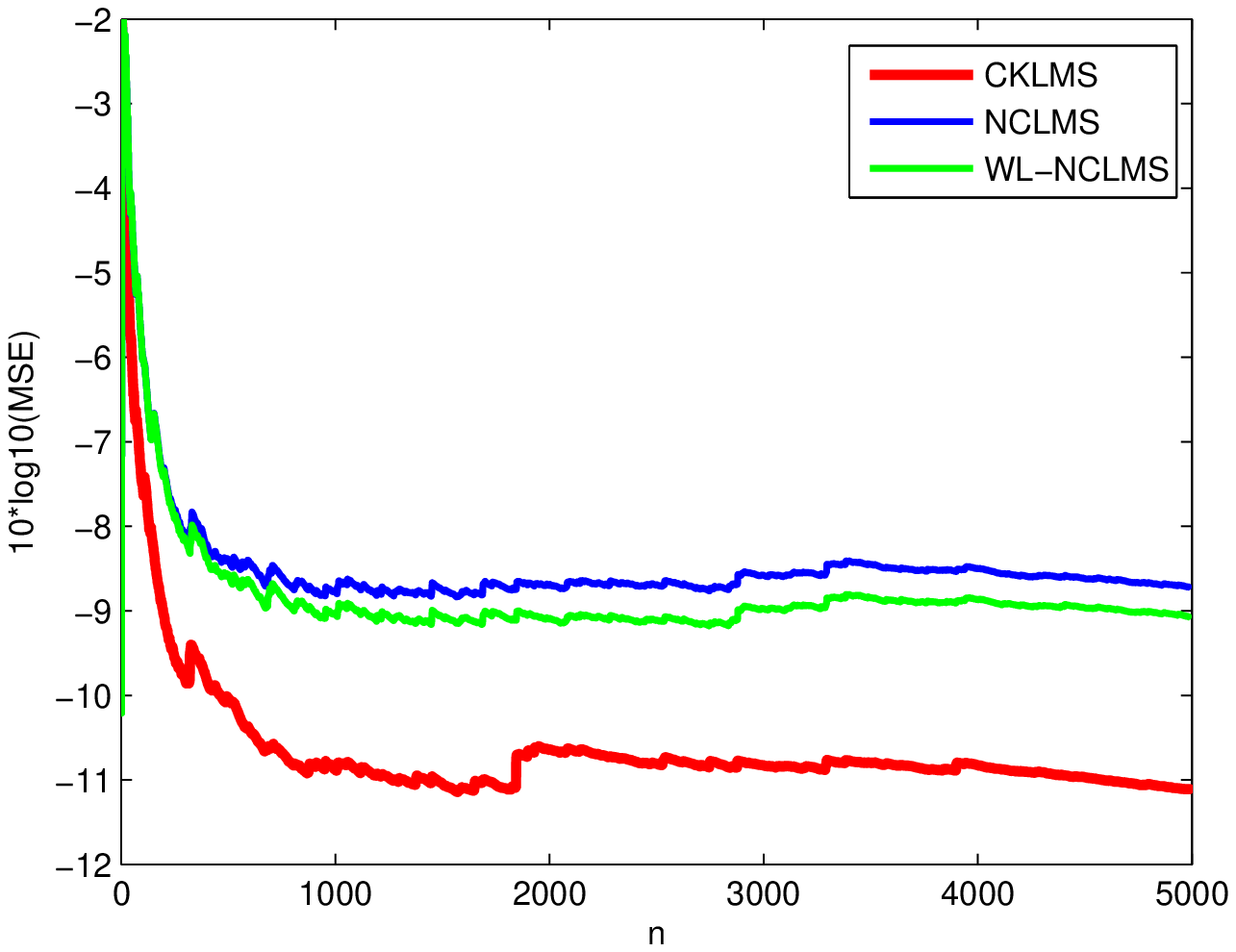}
\centerline{(a)\hspace{12em}(b)}
\end{center}
\caption{Learning curves for KÍCLMS, ($\mu=1$) ÍCLMS ($\mu=1/16$) and WL-ÍCLMS ($\mu=1/16$) (filter length $L=5$, delay $D=2$) in the nonlinear channel equalization, for the (a) circular input case and (b) the non-circular input case.}\label{FIG:exper}
\end{figure}

Experiments were conducted on a set of 5000 samples of the input signal (\ref{EQ:input}) considering both the circular and the non-circular case.
The results are compared with the NCLMS and the WL-NCLMS algorithms. In all algorithms the step update parameter $\mu$ is tuned for best possible results. Time delay $D$ was also set for optimality. Figure \ref{FIG:exper} shows the learning curves of the NCKLMS with $\sigma=5$, compared with the NCLMS and the WL-NCLMS algorithms. Novelty criterion was applied to the CKLMS for sparsification with $\delta_1=0.1$ and $\delta_2=0.2$. In both examples, CKLMS considerably outperforms both the NCLMS and the WL-NCLMS algorithms. However, this enhanced behavior comes at a price in computational complexity, since the CKLMS requires the evaluation  of the kernel function on a growing number of training examples.

\bibliographystyle{IEEEbib}
\bibliography{refs}

\end{document}